# Sentiment Analysis for Roman Urdu Text over Social Media, a Comparative Study

[1] **Irfan Qutab;** [2] **Khawar Iqbal Malik;** [3] **Hira Arooj**

[1] Department of Commuter Science and IT, The University of Lahore
Sargodha, Pakistan

[2] Department of Commuter Science and IT, The University of Lahore
Sargodha, Pakistan

[3] Department of Mathematics and Statistics, The University of Lahore
Sargodha, Pakistan

**Abstract** – In present century, data volume is increasing enormously. The data could be in form for image, text, voice, and video. One factor in this huge growth of data is usage of social media where everyone is posting data on daily basis during chatting, exchanging information, and uploading their personal and official credential. Research of sentiments seeks to uncover abstract knowledge in Published texts in which users communicate their emotions and thoughts about shared content, including blogs, news and social networks. Roman Urdu is the one of most dominant language on social networks in Pakistan and India. Roman Urdu is among the varieties of the world's third largest Urdu language but yet not sufficient work has been done in this language. In this article we addressed the prior concepts and strategies used to examine the sentiment of the roman Urdu text and reported their results as well.

***Keywords*** – L*exicon, Urdu sentiments, Pre-processing, Corpus, Datasets, sentiment classification,*

## 1. Introduction

Generally social media is used to convey sentiment through declaration of opinion or concern about an event. In company, this declaration expressed the company's view and concern to a facility or system [1]. Organizations may use this input to strengthen and maximize their efficiency. Classification is an area where multiple disciplines computing sciences including machine learning, everyday common language, and artificial intelligence converge.

Natural language processing is mostly towards linguistic, although Numerical operations and statistical techniques contribute to text mining. It is common knowledge of phrases and words. People of Pakistan and India normally describe their emotions in Roman Urdu language over social media. Roman Urdu has arisen in especially frequent language in recent years, so people have an option to explore feelings and show their emotions in their own words. Recognized emotion analyzers with particularly well-learned languages, Identical English, stay not feasible created for Urdu or Roman Urdu owing to their variations in writing, morphology, and vocabulary.

People express their reviews while, for example, customers get the products at their doorstep and make its payments online and share their opinions about products and services over social media afterwards. Whenever another person plans to purchase a product, He or she probably wants to view product recommendations of those who bought the same product [1].

Sentiment analysis may be used as a tool to represent the polarity of perceptions and beliefs. It may be stated in different languages, i.e. Arabian, French, Urdu, Hindi, and Urdu among others. Studies concentrated mainly on public polls in the English language.

In the Urdu language or Roman Urdu theoretical practice are quite limited. Sentiment research, commonly known as "opinion mining," describes the Sentiments under a variety of terms to obtain awareness about anonymous representations about opinions, views and feelings.

The study of sentiment shows semantic understanding of written text where people post their views on platforms such as forums and social networks [3]. In Pakistan and India, there's no record of cellular customers who use Roman Urdu for communication. Nowadays other problems such as knowledge retrieval, document start, automated translation, and questions-answer structures are attempted with natural language processing [4].

Roman Urdu has arisen in especially frequent language in recent years, so people have an option to explore feelings and show their emotions in their own words. Hindi / Urdu is the third largest language in the world [5]. Emotions are classified as thoughts and opinions. It is a feeling of somebody that he communicates, either in writing or in spoken way. Emotion may be defined as a positive feeling or a negative one [6].

In this article we have discussed the basic importance of Roman Urdu analysis in section I as introduction. In section II we discuss the literature review about articles in which sentiment analysis has been conducted in past. In Section III we discuss the …..while in section IV we describe our future work about sentiment analysis over Roman Urdu.

## 2. Literature Review

### 2.1 Waikato Environment for Knowledge Analysis (WEKA)

M. Bilal et al uses three classification models the classification of Roman Urdu and English text into two classes positive and negative [7]. A blog was used to gather feedback about both Roman Urdu and English languages. These viewpoints are documented in data files as examples for producing 150 positive and 150 negative views in a training set. Three separate models are implemented on the data set and findings of each model are evaluated. Methodology is shown in figure 1.

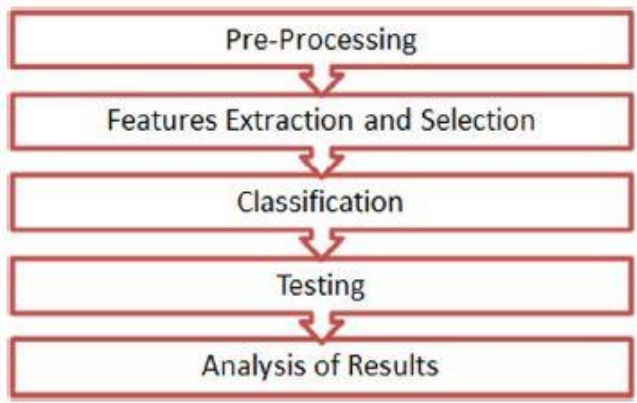

Figure 1: Proposed Model for Roman-Urdu Sentiments[7]

By using this model the results were evaluated. The accuracy, recall, precision, and F1-score generated by Naïve Bays algorithm was greater than the Decision Tree and KNN Classifiers. The comparison of the results of three classifiers is shown in the Figure below.

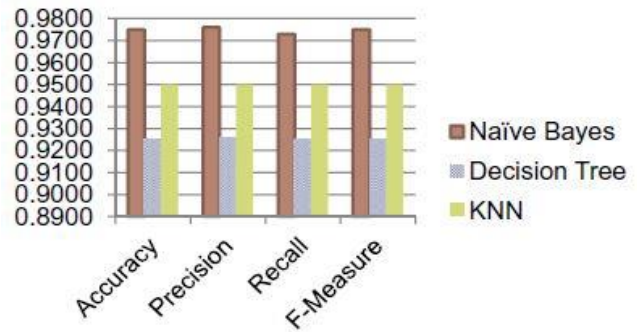

Figure 2: Classification assessment measures results [7]

### 2.2 Support Vector Machine (SVM) for Sentiment analysis of Roman Urdu

F. Noor et, al. discuss the sentiment analysis for Roman Urdu by using SVM classifier. They analyze for three class which were positive, negative and Neutral [6]. Their study focused on Roman Urdu reviews, which are accessible via one of Pakistan's best website Daraz.pk the most popular and viewed websites for e-commerce. A total of 20.286K comments were classified by three experts in three different classes. For extraction of features, which were later transferred to SVM for sentiment classification, the 9.44vector space model and bag of words model was applied. Also MATLAB Linux system experiments were carried out. For future use and research, the dataset is kept available. Basic flow of the framework shown in figure 3.

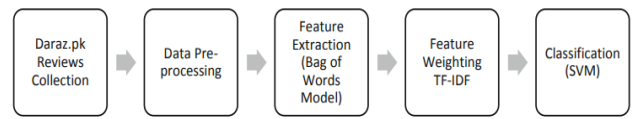

Figure 3: Classifier model for Roman Urdu by using SVM Model [6]

They found the accuracy of 60.03% by using SVM classifier and the results were shown in the form of confusion matrix. They have more than 20 K comments, and the dataset is subsequently split into 2 sets; test sets (20%) and training sets (80%). SVM is used for extraction of the function and classification.

For classification different SVM kernels are used. On a given dataset, the Cubic Kernel achieves the highest accuracy. They achieve the 59.44 % accuracy on training set while with Cubic kernel they achieve 60.03% accuracy on test set.

### 2.3 Recurrent Convolutional Neural Network model for Roman Urdu sentiment analysis

Author in [8] used Recurrent Convolutional Neural Network for sentiments analysis of Roman Urdu text. The purpose of the study was:

- Creation of a labelled data of conventional standards for an under-resourced text classification in Roman Urdu
- Using N-gram, rules-based and RCNN models to evaluate opinion analysis framework.

They perform two types of tests, that is, binary (positive and negative) and tertiary (positive, negative and neutral) classifications. They have evolved high-level system model for identification of the Roman Urdu sentiment. Figure 4 below shows the architecture design.

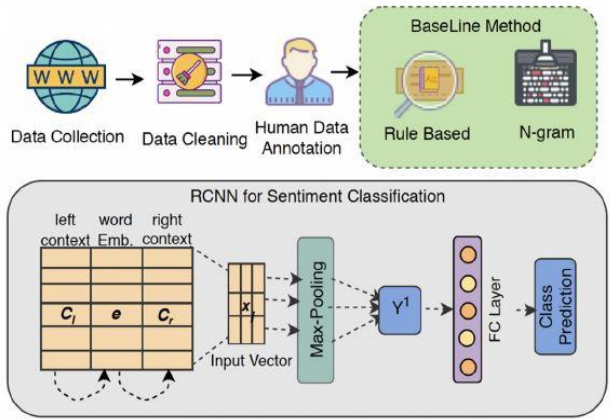

Figure 4: RCNN based classification for Roman Urdu [8]

In the final analysis, the results of the model RCNN are compared with the outputs of the N-grams and Rule-based models.

They found that the RCNN model beats simple models in terms of 0.652 for binary and 0.572 for tertiary classifications, respectively.

They estimate accuracy, Recall, precision and F1 score. The results are shown below.

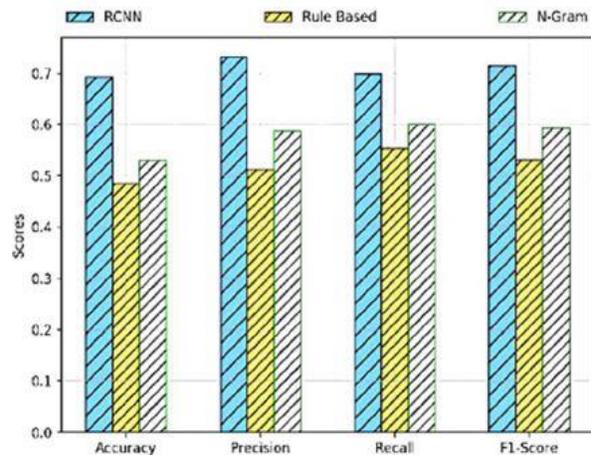

Figure 5: Comparison of results between RCNN, Rule based, and N-Gram models [8]

### 2.4 Hybrid approach with RNN for Roman Urdu sentiment analysis

A Precisely Xtreme-Multi Channel Hybrid Approach was used for Roman Urdu Sentiment Analysis [9]. Because benchmark dataset is not publically available so the Romen Urdu dataset consisting of 3241 sensitivities against positive, negative and neutral classes was used. They adapted various machine learning includes Supporting Vector Machine (SVM), Logistic Regression and Naive Bays. In deep learning they discuss RNN, RCNN and present the hybrid methods to provide benchmark baseline performance on the available data set.

In comparing computer and deeper learning methodologies with 7 and 5 separate feature representation methods, the utility of generated neural word embeddings was evaluated. Finally, it proposes a novel hybrid multi-channel approach that is accurate and incredibly high-tech and provides 9% of adapted machinery and deep learning methods with a 4% F1-score performance. Pretrain terms for Roman Urdu, Word2Vec, Glove, Fast-Text Sensitivity Pretrain word embeddings.

They have implemented an accurately multi-channel technique to harvest the advantages of 3 neural word embedding, using Word2vec on the first [10], Fast-Text on the second channel [11] and Glove on the third channel [12]. Three Gated Recurrent Units (GRUs) have been used on all channels [13]. Central unidirectional GRU retrieves current word, The leftmost and most right-hand unidirectional GRUs are representation by means of the corresponding embedding layer of left and right meaning term. Embeddings are kept static in order to prevent overfitting the model. The representation of three channels using Word2vec, Fast-Text and Glove is combined with every corpus word to create a single representation dependent on semantic similarity. In view of the effectiveness of RNN in long-range addiction learning and the CNN in acquiring promising functionality, a single representation is first moved on to a two-directional GRU to obtain qualitative knowledge better. The most unequal functionality would then be extracted and moved to a fully connected layer through max polling. The proposed model for the system is shown in the figure 6.

In Table 1 performance of adapted methodologies based on machine learning and deep learning is assessed in four assessment metrics by using seven and 5 different representational techniques.

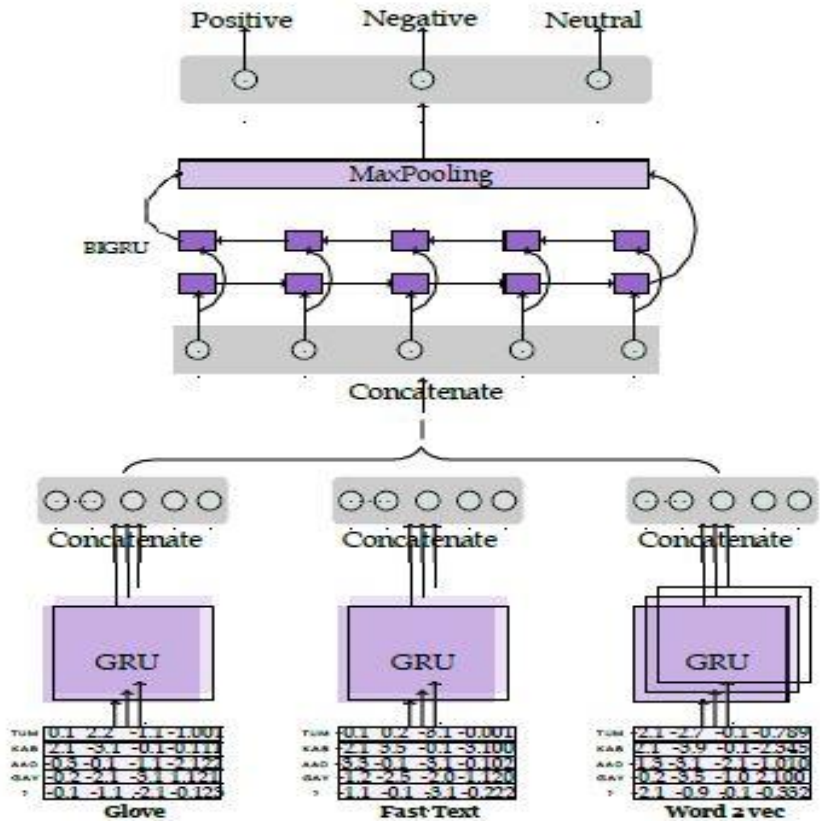

Figure 7: Precisely Extreme Multi-Channel Hybrid Methodology [9]

Table 1: Comparison of deep learning and other adapted methods for hybrid approach [9]

| Algorithms | Embedding | Evaluation Measures | | | |
|---|---|---|---|---|---|
| | | Accuracy | Precision | Recall | F1-score |
| Xtreme multi channel Hybrid Approach | 3(W2V+GloVe+FT) (static), 9 GRU | 0.7708 | 0.7634 | 0.7917 | 0.7581 |
| | 3(W2V+GloVe+FT) (static+ non-static), 9 GRU | 0.8099 | 0.7812 | 0.8136 | 0.7911 |
| | 3(W2V+GloVe+FT) (static), 9 GRU+ Bi-GRU | 0.8243 | 0.8003 | 0.8006 | 0.80 |
| | 3(W2V+GloVe+FT) (static+ non-static), 9 GRU+ Bi-GRU | 0.8417 | 0.8168 | 0.8284 | 0.8221 |

## 2.5 Supervised Learning Hybrid Approach

Another sentiment analysis system was developed for roman Urdu using 3 different supervised machine learning algorithms (i) Logistic Regression with Stochastic Gradient Descent (LRSGD) (ii) Naive Bays (NB) and (iii) Support Vector Machine (SVM) [14].

They focused in this paper on emotion analysis in Roman Urdu's comment / opinions. They create a dataset from various websites containing comments and opinions of Roman Urdu users. By using three Classifiers they concluded that In terms of accuracy, SVM works better than NB and LRSGD. The precision of SVM is obtained at 87.22%. Proposed architecture given by author shown in figure 8

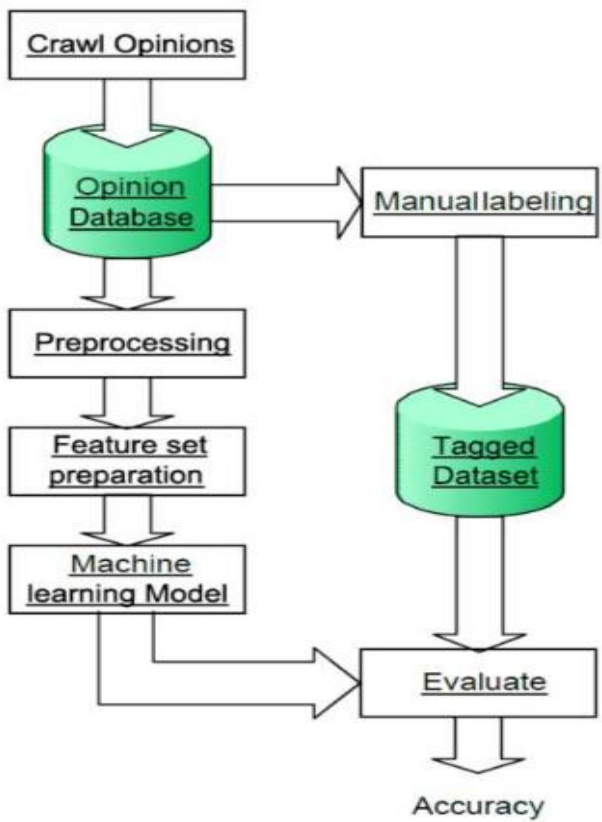

Figure 8: Architecture for Supervised Learning Hybrid Approach

They used Unigram, Bigram and TF-IDF, one R Attribute, Principal component and Gain Ratio Attribute for feature extraction. They combined these techniques to get best features from the dataset, because the individual's results were poor of each method. First they combined (Unigram + Bigram) then (Unigram + Bigram + Tf-idf) similarly (Unigram + Bigram + Tf-idf + Principal Component) and (Unigram + Bigram + Tf-idf +Gain Ratio Attribute). They found the highest accuracy of 87.22%.

They trained three machine learning algorithms NB, LRSGD and SVM for binary classification (to predict emotions of positive or negative opinions) in the dataset. A total of 24 experiments, all three algorithms were applied to the data set by choosing eight feature sets and obtaining results. The tests are done with WEKA, a data mining tool.

In both studies, 10-fold cross validation is used to obtain reliable results. For all eight features, the accuracy of LRSGD is always higher than NB. It was concluded that Unigram+ Bigram + TF-IDF is the best feature set and SVM was the best algorithm for Roman Urdu sentiment analysis.

## 2.6 Sentiment Analysis by using Deep Learning

Deep Learning-Based Sentiment Analysis for Roman Urdu Text was also implemented using Long-short time memory mode (LSTM) and Recurrent Neural Networks (RNN) and three classifiers Naïve bays (NB), Random Forest (RF) and Support Vector Machine (SVM) [15]. They used LSTM because it has tremendous ability to collect data over long-ranges and to solve problems with gradient attenuation and it magnificently displays future meaningful information.

This article is based on applying deep learning techniques to incorporate a study of Roman Urdu emotions.

The empirical results revealed their model 's considerable accuracy, and overcoming the accuracy of the techniques for standard machine learning. RNN efficiency is much higher than traditional lexicon and machine learning methods, as well as statistical techniques like the Hidden Markov (HMM) model.

RNN has an LSTM network [16] which is considered to be a major effort in modeling sequential information such as text and speech.

LSTM is a rapidly growing variant of the RNN model which is commonly used to resolve the gradient of overflow or fading error and to detect long-term dependencies [17][18]. The proposed model is illustrated in the figure below.

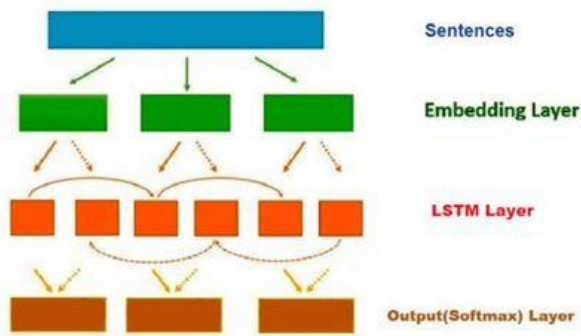

Figure 10: Deep Neural Model, Roman Urdu sentiment analysis [16]

In author's evaluation of the proposed framework of classifier was enforced as a four-layer deep-learning model.

- The input layer (Embedding) is defined for the input layer size is according to the number of inputs. This is equivalent to the scale and number of additional features of a word vector. In order to encrypt any word in the used language, the term vector dimension was fixed to 300.
- Hidden layer: For the hidden LSTM layer, sigmoid activation has been chosen. Input as well as the next layer is completely connected to this layer.
- LSTM layer: In each step t, the input vector x and hidden state h are added to a repeating layer by the recursive operation.
- Output layer: Soft-max activation output layer.

They performed 10-fold cross validation. For this purpose 90% data is used for training and 10% data is used for testing in each fold. They evaluate Precision, Recall, F1-score and Accuracy by using different classifiers. Their experimental evaluations show that the most successful model for sequence data models is the deep neural networks, since no prior knowledge, design and processing methods is needed. Their model has highest accuracy of 95% as compared to other models that were used in this paper.

2.7 Roman Urdu Sentiment Analysis (SA) Hybrid Models

Sentiment analysis of Roman Urdu text by using five different models was conducted [19]. In this paper Roman Urdu data were obtained in 779 reviews in five areas: Drama, Web Reviews, Movie and Tele film, Politics and assorted. They used Bigram, unigram and combined Uni & Bigram features. The accuracy of Logistic Regression model and Naïve bays was better as compared to the other three classifiers. It was also found that after feature reduction, the overall results were better.

The proposed model that has been used for this task is illustrated in the figure below.

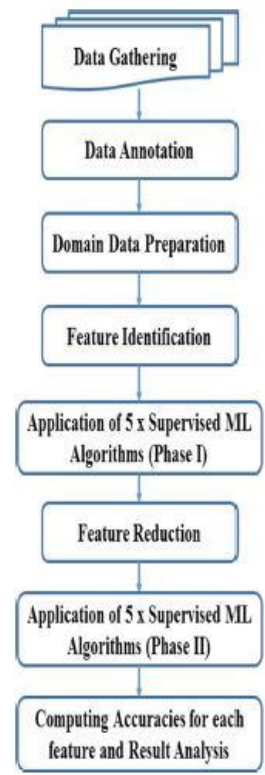

Figure 12: Model used for Roman Urdu Sentiment [19]

According to the methodology after preprocessing they split the data into training and testing by ratio of 60% and 40% after that they identified the features and applied algorithms on these features. In the next step they reduced the features and then again applied the algorithms. Results were much better than the previous results [19].

The comparative analysis by author's for the English and Roman Urdu film reviews is also presented in article and according to that both NB and SVM perform better over features extracted through combined Uni & bigram. In terms of research performance, the best outcomes can be found from the drama and film area. One potential explanation for this is that drama and movies have common aspects such as plots, directors, actors, etc.

## 3. Conclusions

In this paper we have discussed different methods and techniques that were used to achieve the results of sentiment analysis of roman Urdu text/comments. We have tried our best to include all the researches that have been done so far in roman Urdu sentiment classification. In all this study we have seen that classification was performed in two or three classes whether it was (positive, negative or neutral). Most researchers have used Naïve Bays, logistic Regression and Support Vector Machine. We also concluded that very limited work has been done in roman Urdu language. There are few researchers who proposed the models for roman Urdu sentiment analysis. Most of the research is performed in English language. As Roman Urdu is the variant of third largest language in the world there a lot of work is needed in this language

## 4. Future Work

In view of previous studies it is concluded that no work is done on sentiment analysis of roman Urdu by using Multinomial Logistic Regression. In future we will perform sentiment analysis of roman Urdu text by using Multinomial Logistic Regression and will classify the sentiments into multiple classes.

**Authors -**

**Irfan Qutab** is research scholar of MSCS in department of Computer Science and Information Technology at The University of Lahore. He is working over Sentiment Analysis of Roman Urdu as his final research work.

**Khawar Iqbal Malik** has completed MSCS in 2015 and currently student of PhD Computer Science from University of Sargodha. Now he is working as Lecturer in department of Computer Science and Information Technology in The University of Lahore Sargodha campus. His research area is Artificial Neural Network and Information Retrieval Techniques.

**Hira Arooj** has completed MPhil Statistics in 2016 currently she is working as Lecturer statistics in department of Mathematics and Statistics in The University of Lahore Sargodha campus. She is teaching from 4 years in department of computer science. Her research area is statistical Models Applied in NLP & Information retrieval Techniques.